\documentclass[letterpaper]{article} \usepackage{aaai25}  \usepackage{times}  \usepackage{helvet}  \usepackage{courier}  \usepackage[hyphens]{url}  \usepackage{graphicx}    \usepackage{natbib}  \usepackage{caption}       \usepackage{amsmath}
\usepackage{amssymb}
\usepackage{booktabs}
\usepackage{multirow}
\usepackage{hyperref}

\usepackage{algorithm}
\usepackage{algorithmic}

\usepackage{listings}
\DeclareCaptionStyle{ruled}{labelfont=normalfont,labelsep=colon,strut=off} 
\floatstyle{ruled}
\newfloat{listing}{tb}{lst}{}
\floatname{listing}{Listing}
\title{Image Tiling for High-Resolution Reasoning:\\Balancing Local Detail with Global Context}
\author{
Anatole Jacquin de Margerie\textsuperscript{\rm 1}, 
   Alexis Roger\textsuperscript{\rm 2,3},
   Irina Rish\textsuperscript{\rm 2,4}
}
\affiliations{
\textsuperscript{\rm 1}Ecole Polytechnique,
    \textsuperscript{\rm 2}Mila - Quebec AI Institute, 
    \textsuperscript{\rm 3}McGill University,
    \textsuperscript{\rm 4}University of Montreal

}

\begin{document}

\maketitle

\begin{abstract}
Reproducibility remains a cornerstone of scientific progress, yet complex multimodal models often lack transparent implementation details and accessible training infrastructure. In this work, we present a detailed reproduction and critical analysis of the Monkey Vision-Language Model (VLM) \cite{monkey} published in CVPR24, a recent approach to high-resolution image understanding via image tiling. The original paper proposed splitting large images into tiles to recover fine-grained visual details while maintaining computational efficiency.

Our study replicates this strategy using open checkpoints and reimplements the training pipeline. We confirm the key finding of the original Monkey VLM work, namely that tiling effectively recovers local details. We then extend this work further, by investigating the effect of the inclusion of the global context, which provide practical insights for future high-resolution multimodal modeling. However, we also report deviations in the results, with the magnitude of these effects depending heavily on task type and tile granularity.
\end{abstract}

\section{Introduction}
Large Multimodal Models (LMMs) have rapidly advanced the state of the art in tasks that require understanding across both vision and language. Their success relies heavily on the ability of vision encoders to provide meaningful representations of the input images. However, most encoders are constrained to fixed input resolutions, which forces high-resolution images to be resized, often at the cost of the fine details. This limitation is especially problematic in domains such as medical imaging, document understanding, or remote sensing, where both local details and global context are essential.

A straightforward solution is to increase the resolution supported by vision encoders, but this approach comes with steep computational costs and is not always feasible. 
The Monkey VLM framework \cite{monkey} proposed an elegant solution: split large images into tiles, process them independently, and fuse their representations, thereby retaining local detail without prohibitive computational cost.

Despite its conceptual clarity, the Monkey approach raised important questions:
\begin{itemize}
    \item How robust is this technique across different downstream tasks?
    \item To what extent does the inclusion of a resized global view restore lost coherence?
    \item And crucially, can independent researchers reproduce the reported performance under realistic compute budgets?
\end{itemize}

We address these questions by conducting a faithful reproduction of Monkey’s tiling-based visual pipeline within the LLaVA architecture \cite{liu2023improved}, following the original configuration as closely as possible. Our study not only verifies the feasibility of this method but also uncovers task-dependent performance characteristics and implementation nuances that affect reproducibility and applicability. 

Our contributions are:
\begin{enumerate}
    \item A full reproduction of the Monkey VLM training pipeline, including tiling configurations and combined global context inputs.
    \item A systematic evaluation across benchmarks, analyzing trade-offs between local detail and global coherence.
    \item A reflection on the reproduction process, highlighting the task dependant nature of these results. 
\end{enumerate}

\begin{figure}[hb]
\centering
  \centering
  \includegraphics[width=\linewidth]{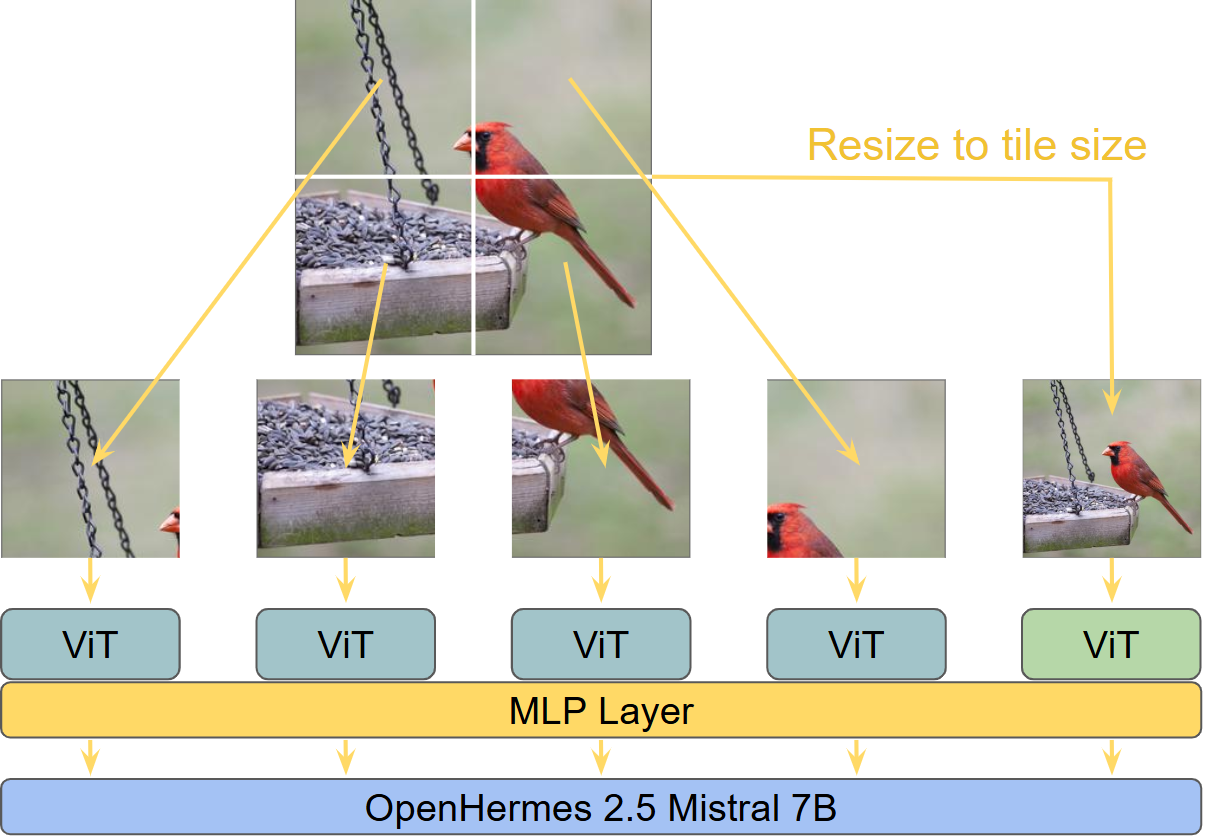}
  \caption{The image splitting approach processes high-resolution images through multiple vision encoder towers, with optional global context.}
  \label{fig:method}
\end{figure}

\section{Related Work}

Our work builds upon recent advances in high-resolution multimodal processing. The LLaVA framework \cite{liu2023improved} established strong baselines for vision-language models but limits the input resolution to what the largest encoder can process. Qwen-VL \cite{bai2023qwen} and PaLI series \cite{chen2023pali,chen2023palix} addressed this through extensive retraining, while Monkey \cite{monkey} proposed the splitting approach we followed and analyzed extensively in this paper.

The Monkey model \cite{monkey} introduced image tiling—splitting an image into non-overlapping tiles (e.g., 2×2, 3×3 grids) and encoding each tile separately. The authors argued this method enables high-resolution reasoning without retraining larger encoders. Moreover, combining these tiles with a downsampled global view purportedly restores spatial coherence.

Reproducing large multimodal models remains difficult due to hidden preprocessing steps, undocumented hyperparameters, and opaque data handling. Prior reproducibility studies in multimodal AI emphasize the importance of transparent design choices and public infrastructure. Our work contributes to this line of inquiry by assessing the faithful reproducibility of Monkey’s tiling paradigm and providing an open reimplementations.

\section{Methodology}

We reimplemented Monkey’s image tiling approach within the LLaVA framework \cite{liu2023improved}, using a Vision Transformer (ViT) \cite{dosovitskiy2020image} (specifically a ViT-S/16 model pre-trained with SigLIP \cite{zhiyuan2023siglip}) as the vision encoder and OpenHermes-2.5-Mistral-7B \cite{jiang2023mistral} as the backbone language model. Although this deviates from the QwenVL model \cite{bai2023qwen} used in the original paper, both models are of the same size and similar performance.

We evaluate 2×2 and 3×3 tiling configurations, with and without global context (a resized version of the full image), using a standard 1×1 split (single resized image) as our baseline. The processing pipeline involves: dividing the image into non-overlapping tiles, resizing the global context, processing all views through separate encoder instances, and concatenating the visual tokens before the projection layer and the language model, as illustrated in Figure~\ref{fig:method}.
Each tile and (optionally) the global image are encoded independently, and their embeddings are concatenated before being projected into the language model space via a shared MLP. No cross-tile attention was used, mirroring Monkey’s design.

Training follows a two-phase approach. First, we pretrain the MLP projection layer on the BLIP-LAION-CC-SBU-558k \cite{blip} dataset with a learning rate of 1e-3 while keeping the vision encoder and language model frozen. Second, we finetune using the \textbf{Monkey training data} \cite{monkey} (summarized in Table~\ref{tab:data}), with a learning rate of 2e-5 and LoRA for the LLM \cite{hu2021lora}. During this stage, the parameters of the vision encoder, the MLP projection layer, and the LoRA adapters applied to the language model are all updated. This mirrors the training setup of the original paper.

\begin{table}[h]
\begin{links}
\centering
  \small
  \captionof{table}{Computational Costs (Training Energy in kWh)}
  \begin{tabular}{@{}ccccc@{}}
  \toprule
  \textbf{Split} & \textbf{Global} & \textbf{Pretrain} & \textbf{Finetune} & \textbf{Model} \\
  \midrule

  \multicolumn{2}{c}{Full img (1x1)} & 22.28 & 50.84 & \href{https://huggingface.co/cerc-aai/mistral-7b-oh-siglip-so400m-finetune-lora}{[Link]} \\
\midrule
2x2 & No  & 22.52 & 53.70 & \href{https://huggingface.co/Anatolejdm/split_2x2_ViT_SO400M_14_SigLIP_384}{[Link]} \\
2x2 & Yes & \textbf{23.97} & 59.08 & \href{https://huggingface.co/Anatolejdm/full_split_2x2_ViT_SO400M_14_SigLIP_384}{[Link]} \\
3x3 & No  & 22.89 & 56.72 & \href{https://huggingface.co/Anatolejdm/split_3x3_ViT_SO400M_14_SigLIP_384}{[Link]} \\
3x3 & Yes & 23.04 & \textbf{63.40} & \href{https://huggingface.co/Anatolejdm/full_split_3x3_ViT_SO400M_14_SigLIP_384}{[Link]} \\

  \bottomrule
  \end{tabular}
  \label{tab:costs}
\end{links}
\end{table}

\section{Experiments}

\subsection{Evaluation Benchmarks}

We evaluated our approach on six established benchmarks covering different aspects of multimodal understanding, from visual reasoning to object hallucination. We used ScienceQA Image \cite{lu2022learn}, GQA \cite{hudson2019gqa}, TextVQA \cite{singh2019towards}, MMVET \cite{yu2023mmvet}, LLaVA-Bench \cite{liu2023visual}, and POPE \cite{Li2023ObjectHallucination}. A detailed description of each benchmark's evaluation format and focus is provided in Appendix.

\subsection{Results}

\begin{table*}[ht]
  \centering
  \caption{Benchmark performance comparison. Results show accuracy (\%) except for the POPE benchmark, where we show the average F1 score over the Random, Popular, and Adversarial Datasets. The first line is the baseline, so with no tiles. The best performing non-baseline result for each benchmark is highlighted in bold.}
  \label{tab:benchmark-results}
  \begin{tabular}{@{}cccccccc@{}}
    \toprule
    \textbf{Split} & \textbf{Global} & \textbf{SQA (Img)} & \textbf{GQA} & \textbf{TextVQA} & \textbf{MMVET} & \textbf{LLaVA Bench} &\textbf{POPE (F1)} \\
    \midrule
    \multicolumn{2}{c}{Full img (1x1)} & 71.34 & 54.08 & 52.99 & 28.30 & 58.40 & 78.57 \\
\midrule
    2x2 & No & 73.53 & 50.80 & 39.83 & 23.20 & 28.60 & \textbf{81.86} \\
    2x2 & Yes & 78.19 & \textbf{54.20} & \textbf{49.96} & 27.10 & \textbf{30.60} & 81.53 \\
    3x3 & No & 66.09 & 38.54 & 36.25 & 15.60 & 16.70 & 33.01 \\
    3x3 & Yes & \textbf{80.17} & 53.82 & 49.63 & \textbf{29.50} & 29.80 & 77.00 \\
    \bottomrule
  \end{tabular}\end{table*}

Table \ref{tab:benchmark-results} recaps the performance of our different models on these benchmarks and reveals several important patterns about the trade-offs between detail preservation and contextual coherence:

\textbf{Detail-Oriented Tasks:} On ScienceQA-Image, high-resolution tiling alone provided a benefit, with the 2×2 split (73.53\%) outperforming the baseline (71.34\%). However, performance was maximized by combining local detail with global context, as the 3×3 split with global view achieved 80.17\%—significantly surpassing all other configurations. This demonstrates that while tiling recovers fine details, the global view provides essential contextual scaffolding to interpret them correctly.

\textbf{Coherence-Sensitive Tasks:} On GQA, which measures spatial and semantic reasoning, the 2×2 configuration with global context (54.20\%) essentially matched baseline performance (54.08\%), showing that global context can fully mitigate the coherence loss from splitting.

\textbf{Challenging Fusion Tasks:} On TextVQA and MMVet, which require integrating information from multiple visual elements, results were mixed but revealing. For MMVet, the 3×3 split with global context (29.50\%) outperformed the baseline (28.30\%), demonstrating that high-resolution tiling can enhance performance on complex, multi-capability benchmarks. However, on TextVQA, all tiling configurations still lagged behind the baseline, suggesting that while our approach helps with general visual reasoning, the simple concatenation method may be suboptimal for tasks specifically requiring OCR and reasoning about text scattered across different image regions.

\textbf{Conversational Tasks:} On LLaVA-Bench, which evaluates open-ended response quality, all tiling configurations performed much worse than the baseline. This suggests that the fragmented visual representation created by concatenation hinders the model's ability to produce coherent, comprehensive, narratives about images.

\textbf{Hallucination Evaluation:} On POPE, which measures object hallucination, the 2×2 configurations showed strong performance, exceeding the baseline. This suggests that access to higher-resolution local patches may provide more grounded visual information, reducing the model's tendency to hallucinate objects. However, the 3×3 split without global context performed poorly, indicating that extreme fragmentation without contextual guidance can be detrimental.

\subsection{Computational Efficiency}

As shown in Table \ref{tab:costs}, our approach maintained computational efficiency throughout training. The addition of global context incurred modest cost increases (5-12\% during finetuning), while the different splitting configurations showed minimal pretraining cost differences. This efficiency is in stark contrast to the resource-intensive full-model retraining approaches used by other high-resolution methods. For instance, training Qwen-VL \cite{bai2023qwen} required thousands of GPU days, representing an energy cost orders of magnitude higher than our most expensive configuration. By building on top of open off-the-shelf components, the method provides a practical path to high-resolution capability without prohibitive computational demands.

\section{Insights from the Reproduction}
\subsection{Confirmed Findings}
Our reproduction successfully recovered the key qualitative trends originally reported in the Monkey paper. First, we confirmed that image tiling substantially improves the model’s ability to recognize and reason about fine-grained visual details. Tasks that rely on localized features, such as small text or subtle object boundaries, benefited most from this strategy. Second, we verified that incorporating a global view alongside the tiled inputs effectively mitigates the coherence loss that occurs when images are processed purely as disjoint patches. This combination enables the model to better integrate spatial relationships and maintain a consistent understanding of the overall scene. Finally, we observed that the computational overhead introduced by tiling remains modest. Even when multiple tiles and a global image are processed, the total energy and compute cost stay orders of magnitude lower than approaches requiring full-model retraining for higher resolutions.

\subsection{Deviations and Implementation Challenges}
While we were able to reproduce most of the Monkey paper’s results, several aspects required special consideration or clarification. The original publication presented its findings as uniformly positive across all benchmarks, yet our experiments revealed a more nuanced reality: optimal performance on one benchmark did not always translate to superior results on others. Instead, we found that tiling effectiveness is task-dependent, and that the ideal grid configuration varies according to the nature of the visual challenge—detail-oriented tasks favored finer splits, while global reasoning tasks required fewer, larger tiles or the inclusion of a global context.

We also encountered several under-specified implementation details that affected reproducibility. These included the precise ordering of patch embeddings, the data preprocessing pipeline, and hyperparameter settings for fine-tuning. These details could lead to measurable performance differences. This underscores the importance of explicit documentation and open training pipelines for reproducibility in multimodal research.

\subsection{Lessons Learned}
From this reproduction effort, we derived several broader lessons relevant to the reproducibility of multimodal systems. First, explicit patch handling and normalization details must be carefully documented to ensure consistent results across reproductions. Small implementation choices, such as how patches are normalized or ordered, can meaningfully alter outcomes. Second, the fusion design between local and global representations plays a critical role. While we replicated the simple concatenation approach used in the original work, we note that more sophisticated attention-based fusion mechanisms could further improve coherence, an avenue worth exploring in future research. Third, transparent data curation and preprocessing are as vital as architectural design choices; undocumented dataset filters or augmentations can easily confound replication attempts.

Finally, our results reaffirm the central importance of global context in maintaining semantic coherence. Across all benchmarks, including a resized global view enabled improved performance, even on tasks where naive tiling alone degraded accuracy. This finding highlights the necessity of balancing fine-grained detail with holistic scene understanding in future high-resolution multimodal architectures.

\section{Discussion}
The tiling+global strategy remains a promising, resource-efficient way to extend LMMs to high-resolution imagery. However, our results demonstrate a clear trade-off: image splitting preserves fine details but disrupts global coherence. The effectiveness of this approach depends critically on the task type, the integration strategy and the available resources.

\subsection{The Critical Role of Global Context}

The consistent improvement from adding global context to the splits across all benchmarks confirms its importance as a spatial scaffold for interpreting tile relationships. This aligns with human visual processing, where both focal detail analysis and global contextual integration are essential \cite{karpathy2015deep}. Our findings suggest that effective high-resolution processing in LMMs requires a similar dual pathway.

\subsection{Task-Dependent Effectiveness}

The variation in results reveals that tiling is task-dependent: it excels at detail-oriented tasks (ScienceQA) where high-resolution processing is crucial, but struggles with narrative coherence (LLaVA-Bench) where our fusion method proves insufficient for holistic understanding. This performance trade-off comes at a modest computational cost, making our approach an efficient strategy for applications requiring fine-grained visual analysis.

\subsection{Future Directions: Toward Adaptive Splitting}

The task-dependent nature of our results points toward adaptive splitting strategies as the most promising future direction. Rather than fixed grids, models could dynamically adjust their processing strategy based on image content and task requirements.

In specialized domains such as medical imaging, different vision towers could be optimized for specific regions of interest (e.g., a dedicated heart split in a full chest scan). Content-aware tiling could apply finer splits to detailed regions while using coarser processing for homogeneous areas. For domain-specific models, learning-based attention mechanisms could identify regions requiring high-resolution analysis, mirroring human foveal vision \cite{karpathy2015deep}.

These adaptive approaches would better balance the detail-coherence trade-off, potentially overcoming the current limitations in generative tasks while preserving the significant gains achieved on detail-oriented benchmarks.

\section{Conclusion}
We reproduced and critically analyzed the Monkey VLM framework \cite{monkey} for high-resolution visual reasoning. Our reproduction validates key claims, especially regarding tiling enhances local detail, and global context restores coherence, while revealing new nuances about task sensitivity and implementation fragility.

We advocate that future works provide complete training scripts and preprocessing pipelines, not just configuration summaries, as even minor implementation gaps significantly hinder reproducibility.

We thank the authors of Monkey VLM for their high quality contributions and paper that made this reproduction work possible.

\bibliography{main}

\newpage
\onecolumn
\section{Appendix}

\subsection{Code}
\begin{links}
    \link{The code can be found in the following repository}{https://github.com/Anatole-JDM/Image_splitting_methods}
\end{links}

\subsection{Evaluation Benchmarks}

We evaluated our approach on six established benchmarks covering different aspects of multimodal understanding, from visual reasoning to object hallucination. Each benchmark employs a specific evaluation protocol:

\begin{itemize}
\item \textbf{ScienceQA (Image)} \cite{lu2022learn}: A multiple-choice question-answering benchmark that tests scientific knowledge and reasoning with images. Performance is measured by accuracy.

\item \textbf{GQA} \cite{hudson2019gqa}: A question-answering benchmark for real-world visual reasoning and compositional question answering. It features balanced distributions of question types and requires understanding of spatial relationships and object attributes. Evaluation is based on accuracy.

\item \textbf{TextVQA} \cite{singh2019towards}: Evaluates reading comprehension in images by requiring models to answer questions about text present in visual scenes. This open-ended generation task is evaluated using average accuracy, requiring models to generate exact matches to ground truth answers.

\item \textbf{MMVET} \cite{yu2023mmvet}: A comprehensive multimodal benchmark evaluating capabilities across perception, knowledge, reasoning, and complex tasks like OCR and detailed description. It uses GPT-4 to evaluate model responses on a 0-5 scale for quality and accuracy.

\item \textbf{LLaVA-Bench} \cite{liu2023visual}: Assesses performance on real-world user queries through human preference evaluation. GPT-4 is used to compare model responses against ground truth answers across three categories: conversation, detail description, and complex reasoning.

\item \textbf{POPE} \cite{Li2023ObjectHallucination}: (Prompt-based Object Probing Evaluation) specifically measures object hallucination in LVMs. It presents models with binary questions (e.g., "Is there a \textit{object} in the image?") across three carefully constructed question types: \textit{random} (unlikely objects), \textit{popular} (plausible but absent objects), and \textit{adversarial} (designed to exploit model biases). Performance is evaluated using precision, recall, and F1 score for this binary classification task.
\end{itemize}

\vspace{5mm}
\subsection{Monkey Training Data}
\label{app:monkeydata}

\begin{table*}[ht]
\centering
\begin{tabular}{c|c|c}
\toprule
\textbf{Task} & \textbf{Dataset} & \textbf{Samples} \\
\midrule
\multirow{3}{*}{Image Caption} 
    & Detailed Caption & 213k \\
    & COCO Caption~\cite{karpathy2015coco} & 82k \\
    & TextCaps~\cite{sidorov2020textcaps} & 109k \\
\midrule
\multirow{5}{*}{General VQA} 
    & VQAv2~\cite{goyal2017making} & 100k \\
    & OKVQA~\cite{marino2019ok} & 18k \\
    & GQA~\cite{hudson2019gqa} & 150k \\
    & ScienceQA~\cite{lu2022learn} & 18k \\
    & VizWiz~\cite{gurari2018vizwiz} & 20k \\
\midrule
\multirow{3}{*}{Scene Text VQA} 
    & TextVQA~\cite{singh2019towards} & 34k \\
    & OCRVQA~\cite{mishra2019ocr} & 250k \\
    & AI2D~\cite{kembhavi2016diagram} & 24k \\
\midrule
\multirow{8}{*}{Doc-oriented VQA} 
    & DocVQA~\cite{mathew2021docvqa} & 118k \\
    & ChartQA~\cite{masry2022chartqa} & 84k \\
    & InfoVQA~\cite{mathew2022infographicvqa} & 47k \\
    & DeepForm~\cite{deepform} & 7k \\
    & KLC~\cite{stanislawek2021kleister} & 27k \\
    & WTQ~\cite{pasupat2015compositional} & 28k \\
    & TabFact~\cite{chen2019tabfact} & 91k \\
    & VisualMRC~\cite{tanaka2021visualmrc} & 21k \\
\midrule
\textbf{Total} & -- & \textbf{1.44m} \\
\bottomrule
\end{tabular}
\caption{Details on the Monkey training data, derived entirely from publicly available datasets.}
\label{tab:data}
\end{table*}

\end{document}